\pdfoutput=1
\documentclass[10pt,twocolumn]{article}

\usepackage{conf}
\pdfoutput=1
\usepackage{times}
\usepackage{graphicx}
\usepackage{here}
\usepackage{amsmath}
\usepackage{amssymb}
\usepackage{comment}
\usepackage{mathtools}

\newcommand{\equref}[1]{{eq \ref{#1}}}
\newcommand{\figref}[1]{{Figure \ref{#1}}}
\newcommand{\tabref}[1]{{Table \ref{#1}}}

\newcommand{\argmin}{\mathop{\rm arg~min}\limits}
\newcommand{\twolines}[2]{\begin{tabular}{c}#1 \\ #2\end{tabular}}
\usepackage[pagebackref=true,breaklinks=true,letterpaper=true,colorlinks,bookmarks=false]{hyperref}

\conffinalcopy 

\ifconffinal\fi

\begin{document}

\title{Efficient training for future video generation based on \\ hierarchical disentangled representation of latent variables}

\author{Naoya Fushishita\\
The University of Tokyo\\
{\tt\small fushishita@mi.t.u-tokyo.ac.jp}
\and
Antonio Tejero-de-Pablos\\
The University of Tokyo\\
{\tt\small antonio-t@mi.t.u-tokyo.ac.jp}
\and
Yusuke Mukuta\\
The University of Tokyo / RIKEN\\
{\tt\small mukuta@mi.t.u-tokyo.ac.jp}
\and
Tatsuya Harada\\
The University of Tokyo / RIKEN\\
{\tt\small harada@mi.t.u-tokyo.ac.jp}
}

\maketitle
\ifconffinal\thispagestyle{empty}\fi

\begin{abstract}
    Generating videos predicting the future of a given sequence has been an area of active research in recent years. However, an essential problem remains unsolved: most of the methods require large computational cost and memory usage for training. In this paper, we propose a novel method for generating future prediction videos with less memory usage than the conventional methods. This is a critical \textit{stepping stone} in the path towards generating videos with high image quality, similar to that of generated images in the latest works in the field of image generation. We achieve high-efficiency by training our method in two stages: (1) image reconstruction to encode video frames into latent variables, and (2) latent variable prediction to generate the future sequence. Our method achieves an efficient compression of video into low-dimensional latent variables by decomposing each frame according to its hierarchical structure. That is, we consider that video can be separated into background and foreground objects, and that each object holds time-varying and time-independent information independently. Our experiments show that the proposed method can efficiently generate future prediction videos, even for complex datasets that cannot be handled by previous methods.
\end{abstract}

\section{Introduction}
In the field of generative modeling, one of the tasks that is gaining major attention is video generation, in particular future prediction video generation.
This task involves predicting and generating a sequence of future frames given an input video. It has a wide range of applications: from self-driving cars to sports video analysis, and assisting in video production of animation and movies; and it is being researched as actively as the task of image generation \cite{vondrick2016generating, denton2017unsupervised, tulyakov2018mocogan, mathieu2015deep, finn2016unsupervised, van2017transformation, jia2016dynamic, villegas2017learning, walker2017pose, cai2018deep, fushishita2019long, villegas2017decomposing, hsieh2018learning}.
Compared to image generation, video generation involves a larger number of dimensions, which in turn requires a larger computational cost and memory usage. Therefore, research on frameworks that can generate videos at low cost is indispensable for the progress in the field. For example, while image generation methods can successfully achieve a resolution of $1024 \times 1024$ pixels~\cite{Karras2019stylegan2}, future video generation methods are not able to reach high quality. Besides, as long as an efficient framework is not established, the application of future video generation will be restricted to a few people with high-performance GPUs.
However, as far as we know, not many studies approach this problem~\cite{TGAN2020, clark2019efficient}.

In this paper, we propose a novel method that can efficiently generate future prediction videos while using less memory than conventional video generation methods.
We leverage the hierarchical structure of videos, that is, video decomposition into background and foreground, while the foreground itself consists of several objects. Each object in the foreground independently stores pose information that changes with time and content information that remains constant within the video. For example, if the foreground object is a human, then its location, posture and facial expressions are pose information that changes with time, while its clothes, skin color and the face itself are content information that does not change with time.
Our method encodes each video frame into disentangled latent variables that are decomposed in its hierarchical structure.
To generate natural videos, existing methods have considered pose and content information separation~\cite{villegas2017decomposing, denton2017unsupervised, tulyakov2018mocogan, ohnishi2018hierarchical, hsieh2018learning}, foreground and background separation~\cite{vondrick2016generating, ohnishi2018hierarchical} and so on. However, to the best of our knowledge, this is the first method that considers a fine-grained separation of the video hierarchical structure.
Then, for a given ``past'' sequence of latent variables, we train the latent variable sequence generator to predict the ``future'' latent variables. This two-stage training can predict future video with low memory usage.

Our contributions are as follows:
\begin{itemize}
  \item We propose a novel framework for efficient future video generation that leverages a two-stage structure for training.
  \item For this first stage, we propose a novel method for learning disentangled latent variables according to the hierarchical structure of videos.
  \item We provide an exhaustive evaluation, proving that our videos' performance is more consistent and efficient than conventional methods, even for complex video datasets.
\end{itemize}

\section{Related Works}
\subsection{Video generation and future prediction video generation}

One of the major methods common to both future prediction video generation and unconditional video generation is to first generate a sequence of latent variables corresponding to each video frame, to then generate each video frame~\cite{saito2017temporal, tulyakov2018mocogan, denton2017unsupervised}. Compared methods that generate all video frames at once via 3D CNN~\cite{vondrick2016generating, ohnishi2018hierarchical}, this methodology processes the xy-coordinate (i.e., space) and the t-coordinate (i.e., time) separately, which makes it more suitable to grasp the video structure.

In addition, many prior studies are aimed to improve the quality of the generated videos by using decomposed latent variables.
In particular, in future prediction video generation methods such as DRNET \cite{denton2017unsupervised}, clean decomposition is needed to be consistent with the input video. Without such decomposition, long future videos tend to get blur or collapse at the end.
One major decomposition method is the separation of pose information, which varies with time, and content information, which is time-independent~\cite{villegas2017decomposing, denton2017unsupervised, tulyakov2018mocogan, ohnishi2018hierarchical, hsieh2018learning}. In this approach, the content information is represented by a single content latent variable that is common to all frames, and the temporal variations of the video are represented by pose latent variables that vary among frames.
There are other decomposition methods such as separation of foreground and background~\cite{vondrick2016generating, ohnishi2018hierarchical} and separation of each object~\cite{hsieh2018learning}. However, to the best of our knowledge, no previous work considers multiple decomposition techniques at the same time, such as our video decomposition according to the hierarchical structure proposed in this paper.

\subsection{Video generation with small cost}
Compared to image generation, one of the main challenges of video generation is its large computational cost and memory usage. To approach this, previous works proposed methods that can generate videos at a smaller cost.
In TGANv2 \cite{TGAN2020} and DVDGAN \cite{clark2019efficient}, the computational cost is reduced by reducing the size of the input and the intermediate feature maps.
However, the computational cost is still high compared to image generation methods, as most methods still need to use all video frames at once during training. For this reason, our method avoids using the whole video at once.



\section{Efficient training for future video generation}
\subsection{Overview of the proposed method\label{sec:pipeline}}


      
      


We propose a framework to generate a natural future prediction video as a continuation of a given input video sequence.
We train the video generator in two stages. 
In the first stage, we train an image reconstruction network (VAE~\cite{kingma2013auto}) that encodes each frame of the video into latent variables and reconstructs the original frame. Next, each frame of the video in the dataset is encoded into latent variables of small dimensionality using the encoder of the trained VAE to create a latent variable sequence dataset corresponding to the video dataset. Finally, the resulting latent variable sequence dataset is used to train a latent variable sequence generator that predicts the future latent variable sequence from the given past latent variable sequence. The video is then generated by decoding the predicted future sequence.

This two-stage training is similar to DRNET~\cite{denton2017unsupervised}, but differs significantly in two ways. The first difference is that the author's implementation of DRNET encodes the video into latent variables every iteration in the second training stage, whereas in this study, the video dataset is converted into a latent variable sequence dataset only once between the first and second training stages. This eliminates the need to handle the entire video during training and reduces memory usage. More specifically, the first training stage uses about the same amount of memory as conventional image generation methods, and the second training stage uses less memory than conventional video generation methods because the dimensionality of the latent variables is much smaller than that of an image frame.
The size of the intermediate feature maps in our method is approx.  $\mathcal{O}(BHWC + BTk)$, whereas the size in conventional video generation methods is approx. $\mathcal{O}(BTHWC)$. Here $B$ is the batch size, $H$ and $W$ are the frame's height and width, $T$ is time duration, $C$ is the channel size, and $k$ is the dimension of the latent variables ($k \ll HWC$).

The second difference is the way the latent variables are decomposed. In order to reduce the amount of memory used, the training of the latent sequence generator does not use the video pixel reconstruction error, which makes the learning process more complex than the conventional end-to-end method. In order to facilitate the second training stage, we propose obtaining disentangled latent variables from the VAE trained in the first stage.
Whereas DRNET simply decomposes the latent variables into \textit{pose} and \textit{content} latent variables, our method's decomposition is based on the hierarchical structure of the video. That is, we disentangle the latent variables that represent the background and each object in the foreground, and within the latent variable for the objects, we further disentangle it into (1) latent variables that represent the position of the object, (2) pose latent variables that represent time-varying information about the object, and (3) content latent variables that represent time-invariant information about the object. Such a detailed hierarchical decomposition allows applying our method to complex datasets.


\subsection{Image reconstruction network \label{sec:image_generator}}

\begin{figure}[tbp]
  \begin{center} 
    \includegraphics[width=\linewidth]{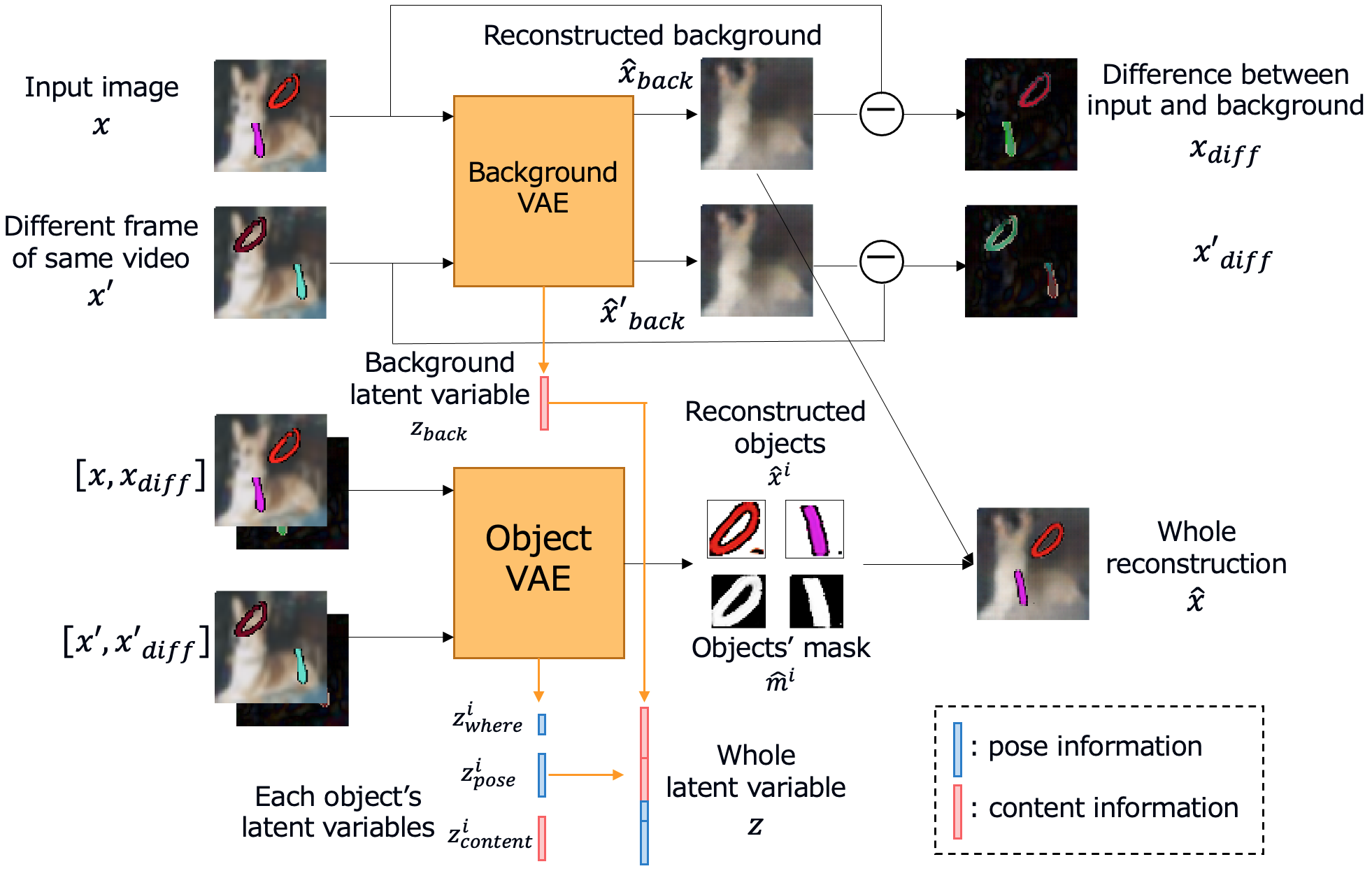}
    \caption{The whole pipeline of our image reconstruction network. It consists of two networks: the background VAE, which reconstructs only the background, and the object VAE, which reconstructs only each foreground object and its mask image.} 
    \label{fig:vae} 
    \vspace{-5mm}
  \end{center}
\end{figure}

This section describes our image reconstruction network, which compresses a video frame into a low-dimensional latent variable and then reconstructs the original frame.
As the main contribution of this work, the image reconstruction network not only reconstructs the original image but also obtains a disentangled hierarchical representation of the structure of the video into the latent variables.

In order to achieve the separation, or disentanglement, of background and foreground, two sub-networks are trained: \textit{background VAE}, which takes a frame as input and reconstructs only the background, and \textit{object VAE}, which reconstructs only the objects in the foreground. Then, the original image is reconstructed by combining the background and objects from each sub-network, and the reconstruction error is calculated.
In order to distinguish between \textit{pose} and \textit{content} information, we use two different randomly selected frames of the same video, $x \in \mathbb{R}^{H \times W \times 3}$ and $x' \in \mathbb{R}^{H \times W \times 3}$, as in DRNET \cite{denton2017unsupervised} for training. Here, $H$ and $W$ are the height and width of the image, respectively (plus 3 RGB channels). We consider the variant features between the two frames as pose information, and the common features as content information.
The overall diagram of the network is shown in \figref{fig:vae}.

\subsubsection{Background VAE}

Background VAE takes a single image as input and encodes only the information about the background in a latent variable, to then reconstruct only the background of the original image from that variable.
Background VAE consists of two networks: an encoder $E_{back}$ and a decoder $D_{back}$.

We encode $x$ and $x'$ independently using $E_{back}$ to obtain the corresponding latent variables $z_{back}$ and ${z'}_{back}$. Then, $z_{back}$ obtained from $x$ and ${z'}_{back}$ obtained from $x'$ are swapped, and decoded using $D_{back}$ to reconstruct the background image. That is, we use the image obtained from ${z'}_{back}$ as the background of the reconstructed $x$. Without this swap, background VAE's reconstruction would include the foreground objects. By swapping, the encoder is trained to extract only information about the parts of the video where the pixel values do not vary among frames. Thus, foreground objects are ignored and the background information can be extracted unsupervisedly.

\subsubsection{Object VAE \label{sec:object_vae}}
\begin{figure}[tbp]
  \begin{center}
    \begin{tabular}{c}
      \begin{minipage}{\linewidth}
        \begin{center}
          \includegraphics[width=\linewidth]{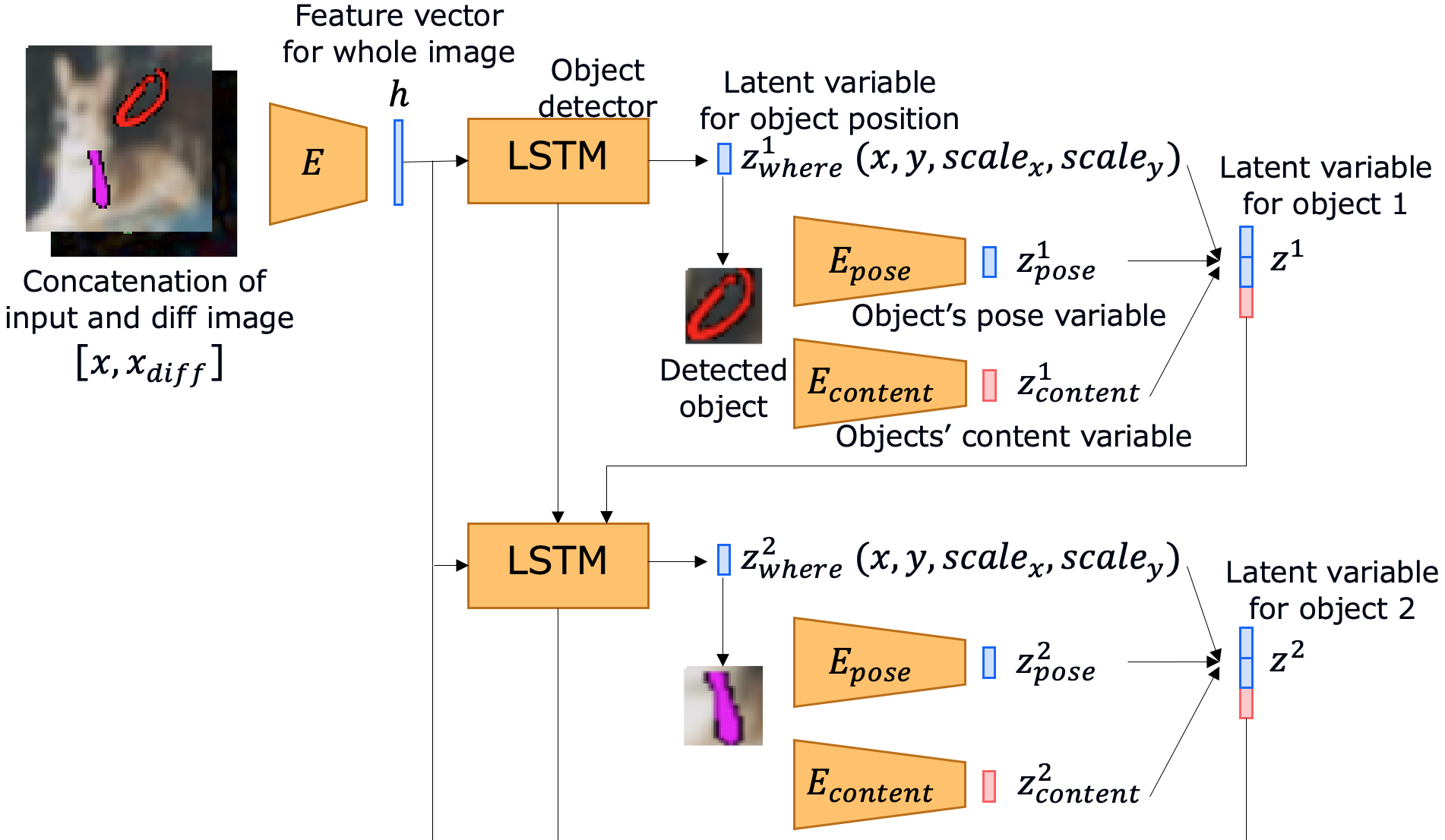}
          \\(a) Inference flow of object VAE
        \end{center}
      \end{minipage} \\

      \begin{minipage}{\linewidth}
        \begin{center}
          \includegraphics[width=\linewidth]{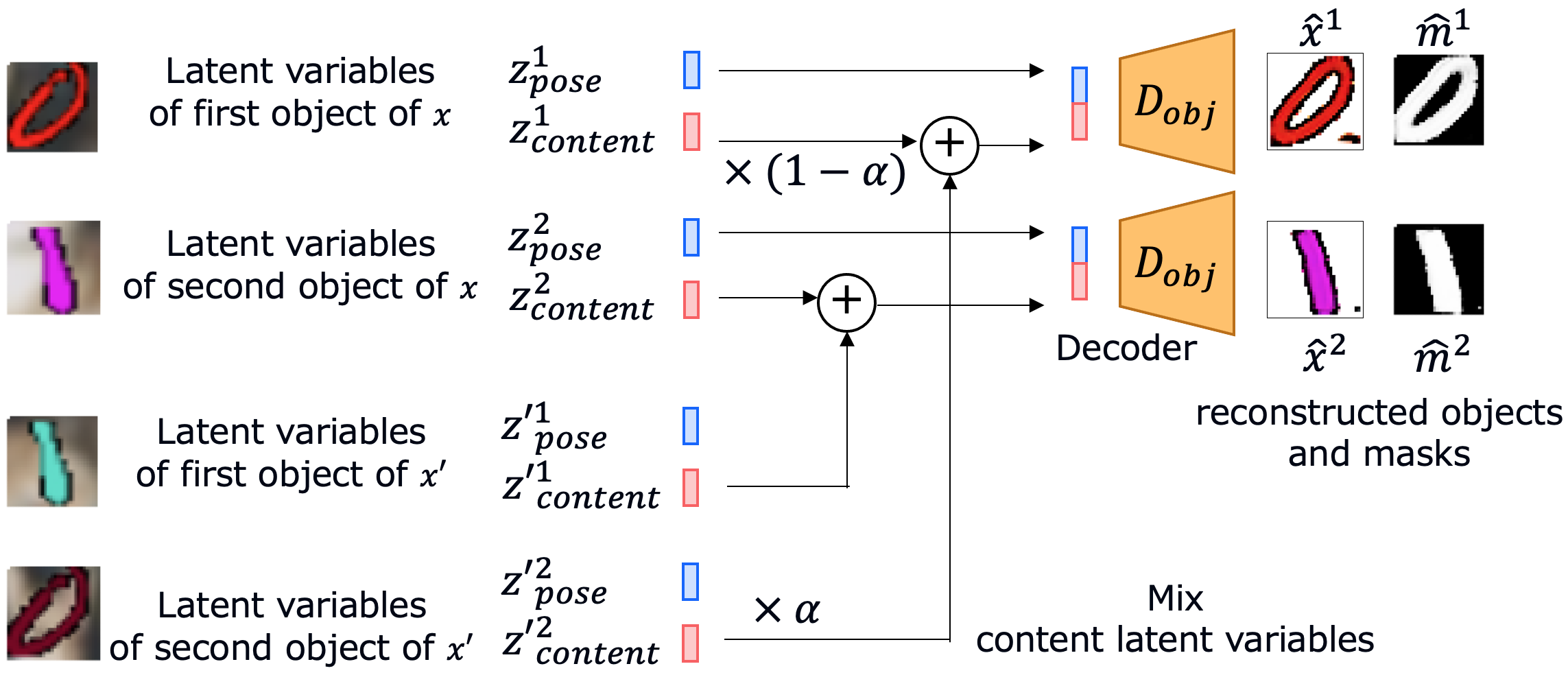}
          \\(b) Generation flow of object VAE
        \end{center}
      \end{minipage}\\
    \end{tabular}
    \caption{An overview of object VAE}
    \label{fig:object_vae}
    \vspace{-5mm}
  \end{center}
\end{figure}

Object VAE detects each object in the image and encodes them into latent variables that contain only the object's \textit{position}, object's \textit{pose information}, and object's \textit{content information}, respectively. Then, it reconstructs each object by decoding the latent variables.

\paragraph{Overview of object VAE}

\figref{fig:object_vae} shows an overview of object VAE. As in background VAE, two different frames of the same video (i.e., $x$ and $x'$) are used simultaneously. This allows the network to disentangle pose and content information for each object by learning the common and variant features between $x$ and $x'$. When the background of $x$ obtained by background VAE is $\hat{x}_{back}$, the difference image between the background and the original images is calculated as $x_{diff} = x - \hat{x}_{back}$. Here, the input of object VAE takes a 6-channel image (3 RGB channels $\times 2$) with $x$ and $x_{diff}$ concatenated in the channel direction. The difference image $x_{diff}$ has pixel values close to $0$ in the background and non-zero values in the foreground. This way, including $x_{diff}$ in the input to object VAE facilitates the detection of each object.

Let $N$ be the maximum number of objects in a video in the dataset, and let $z^i$ be the latent variable corresponding to the $i$-th object $(1 \le i \le N)$ in a frame. In our method, $z_i$ is a concatenation of three types of latent variables, $(z_{where}^i, z_{p}^i, z_{c}^i)$. First, $z_{where}^i$ is a vector that represents the position of the $i$-th object in the image for the input image frame. Specifically, it is a four-dimensional vector that concatenates the $xy$-coordinates of the object in the image and the size of the object in $xy$-directions. Then, $z_{p}^i$ and $z_{c}^i$ are latent variables that contain only the pose and content information of the object, respectively.

\paragraph{Inference of the latent variables from the image}
This section describes the procedure to estimate the $z^i$ of each object from the input image (\figref{fig:object_vae} (a)). Our method is based on AIR~\cite{eslami2016attend}.
First, an image $x$ is encoded using an encoder $E$ consisting of CoordConv~\cite{liu2018intriguing}, which can effectively deal with the coordinates information, to obtain $h$, a feature vector representing the entire image. Next, $h$ is input to an LSTM, the object detector, to estimate the latent variable $z_{where}^1$ for the position of the first object. Then, using the information in $z_{where}^1$, the sub-image containing object $x^1$ is cropped from the original image and resized to a predetermined size using the Spatial Transformer Network \cite{jaderberg2015spatial}. Then, two encoders are used to encode $x^1$, $E_{p}$ and $E_{c}$, resulting in $z_{p}^1$ and $z_{c}^1$. Finally, $z^1$, the latent variable representing the first object, is obtained by concatenating $(z_{where}^1$, $z_{p}^1$ and $z_{c}^1)$. This procedure is repeated $N$ times to obtain the latent variables corresponding to all objects in the image.

Here, in order to disentangle \textit{pose} latent variables from \textit{content} latent variables, we use adversarial learning
to prevent content information from being mixed into the pose latent variables \cite{denton2017unsupervised}.
We introduce a discriminator ${\rm Dis_{p}}$ that takes as input the pose latent variables for two objects and discriminates whether they are obtained from the same object in different frames in the same video or from objects in different videos. 
If the pose latent variables are mixed with content information, the discriminator can detect that they are from the same object by focusing on the content information common to the two pose latent variables. Therefore, in order to fool the discriminator, the encoder is trained not to include content information into the pose latent variables.

\paragraph{Sub-image generation from latent variables}
Next, we describe the procedure for reconstructing sub-images of objects from the obtained latent variables (\figref{fig:object_vae} (b)). For the $i$-th object, from the concatenated latent variable $(z_{p}^i, z_{c}^i)$, a sub-image is generated through the decoder $D_{obj}$ (implemented with the Spatial Broadcast Decoder \cite{watters2019spatial}). This sub-image consists of the $i$-th object $\hat{x}^i \in \mathbb{R}^{h \times w \times 3}$ and a one-channel mask image $\hat{m}^i \in \mathbb{R}^{h \times w \times 1}$ representing whether each pixel belongs to that object or not. Here, $h$ and $w$ are the hyperparameters that represent vertical and horizontal sizes of the sub-images. The mask is used to naturally paste the generated small image onto the background image.
However, we have devised a way to satisfy the requirement that $z_{c}^i$ contains information about only time-independent factors.

Now, we have a latent variable $z^i$ and $z'^i~(1 \le i \le N)$ for each object in $x$ and $x'$. Since $x$ and $x'$ are frames sampled from the same video, we assume that they contain the same objects. Thus, for every $i~(1 \le i \le N)$, there exists $j~(1 \le j \le N)$ such that the $i$-th object in $x$ and the $j$-th object in $x'$ are the same object. Here, the $j$ corresponding to each $i$ can be found by looking for the object with the smallest distance between the content latent variables. So, instead of generating the sub-image containing the $i$-th object of $x$ exclusively from $z_{c}^i$, we generate it from a vector of $z_{c}^i$ and ${z'}_{c}^j$ mixed in random proportions. The above discussion can be summarized in the following equation.
\begin{eqnarray}
    \label{eq:search_nearest}
    j = \argmin_{1 \le j \le N} \|z_{c}^i - {z'}_{c}^j\|^2 \\
    (\hat{x}^i, \hat{m}^i) = D_{obj}(z_{p}^i, (1 - \alpha) z_{c}^i + \alpha {z'}_{c}^j) 
    \label{eq:decoder}
\end{eqnarray}
Here, $\hat{x}^i$ is the generated sub-image for the $i$-th object of $x$ and $\hat{m}^i$ is the small mask image, and $\alpha$ is the ratio of mixing the two content latent variables (the Appendix describes how $\alpha$ is selected).
If the content latent variable contains pose information, the pose information of the $x'$ object will be used to reconstruct the $x$ object, and the reconstruction will not succeed. So, by using mixed content latent variables, we can prevent pose information from being mixed in with the content latent variables.

\subsubsection{Merge of background and objects \label{sec:merge}}
Once the background image $\hat{x}_{back} \in \mathbb{R}^{H \times W \times 3}$ and the reconstructed sub-image $\hat{x}^i \in \mathbb{R}^{h \times w \times 3}$ for each object are obtained from the background and object VAEs, they are combined using $\hat{m}^i \in \mathbb{R}^{h \times w \times 1}$ and $z_{where}^i$.
By leveraging the position information of $z_{where}^i$, we can obtain an object image of the same size as the background image, $\hat{y}^i \in \mathbb{R}^{H \times W \times 3}$, from $\hat{x}^i$, and an image with a mask pasted onto it, $\hat{y}_{m}^i \in \mathbb{R}^{H \times W \times 1}$, from $\hat{m}^i$ using the Spatial Transformer Network \cite{jaderberg2015spatial}. Then, the background and object images are combined as in the following equation, to obtain the reconstructed image of the entire image $\hat{x}\in \mathbb{R}^{H \times W \times 3}$.
\begin{eqnarray}
    \hat{x} = (1 - \sum_{i=1}^N \hat{y}_{m}^i) \odot \hat{x}_{back} + \sum_{i=1}^N \hat{y}_{m}^i \odot \hat{y}^i
    \label{eq:merge}
\end{eqnarray}
Here, $\odot$ is the element-wise product.

Also, by concatenating the latent variable $z_{back}$ obtained from the background VAE and the latent variable $z^i$ for each object obtained from the object VAE, we obtain the latent variable $z$ corresponding to the entire image.

\subsubsection{Objective functions}
This section describes the loss functions used for training the image reconstruction network.

First, as in the original VAE~\cite{kingma2013auto}, the image reconstruction error and the KL divergence between the encoded latent variables and the prior distribution are used as loss functions.
We also used an adversarial loss \cite{goodfellow2014generative, Mescheder2018ICML} for the generated image to be sharp.





Next, we introduce auxiliary loss functions to obtain the disentangled latent variables.
First, since $x$ and $x'$ have a common background, we used \equref{eq:back_sim}, which reduces the difference between $z_{back}$ and $z'_{back}$.
\begin{equation}
    \label{eq:back_sim}
    \mathcal{L}_{back} = \|z_{back} - z'_{back}\|^2
\end{equation}
Similarly, the content latent variables for the same object in different frames should be the same vector. Therefore, we introduced the triplet loss~\cite{schroff2015facenet} for the content latent variable as an auxiliary loss function (\equref{eq:content_sim}).
\begin{eqnarray}
    \mathcal{L}_{c} = [\|z^i_{c} - {z'}^{j_1}_{c}\|^2 - \|z^i_{c} - {z'}^{j_2}_{c}\|^2 + \beta]_+
    \label{eq:content_sim}
\end{eqnarray}
In this equation, $j_1$ represents the object of $x'$ that is closest to the $i$-th object of $x$, and $j_2$ represents the object of $x'$ that is second closest to it. These $j_1$ and $j_2$ can be obtained as \equref{eq:search_nearest}. In other words, by learning to reduce the distance of the content latent variable with the nearest object, and to increase the distance with the second nearest object, we learn to make the content latent variable between the same objects closer.

Also, as described in Sec.~\ref{sec:object_vae}, an adversarial loss is used to prevent content information from being mixed with the pose latent variable. The pose discriminator ${\rm Dis_{p}}$ is trained according to the following loss function.
\begin{equation}
    \label{eq:pose_dis}
    \mathcal{L}_{advD} = - \mathbb{E} [ \log ({\rm Dis_{p}}(z^i_{p}, {z''}^i_{p}))] - \mathbb{E} [ \log (1 - {\rm Dis_{p}}(z^i_{p}, {z'}^j_{p}))]
\end{equation}
Here, ${z''}^i_{p}$ represents the pose latent variable for an object in a different video that is not related to $x$. ${z'}^j_{p}$ represents a latent variable for the object as the $i$-th object in $x$ in $x'$. This $j$ can be estimated in the same way as \equref{eq:search_nearest}.
The pose encoder $E_{p}$ learns to prevent the pose discriminator from identifying two pose latent variables that represent the same object as the same object vectors.
\begin{equation}
    \label{eq:pose_dis_gen}
    \mathcal{L}_{advE} = \mathbb{E} [ \log (1 - {\rm Dis_{p}}(z^i_{p}, {z'}^j_{p}))]
\end{equation}
These are the auxiliary loss functions to facilitate the decomposition of the latent variables.

Finally, we explain the loss function for stabilizing the learning. This network tends to fall into a local minimum where no object is detected by setting all the pixel values of the mask image to $0$. To avoid such a local minimum, we add a loss function \equref{eq:mask_loss} such that the average pixel value of the mask images is about $0.5$ in the first half of the training.
\begin{equation}
    \label{eq:mask_loss}
    \mathcal{L}_{mask} = (\frac{1}{hw} \sum_{y=1}^h \sum_{x=1}^w \hat{m}^i_{y, x} - 0.5) ^ 2
\end{equation}
Once the network training has progressed enough to be able to detect objects, it rarely falls into the aforementioned local minimum. Thus, this loss function is not used in the second half of training, allowing to refine the learned masks.

\subsection{Latent variable sequence generator \label{sec:latent_generator}}

Using the encoders trained in the first stage, the video dataset can be transformed into the corresponding latent variable sequence dataset. The details of the transformation method are given in the Appendix. This section describes the training of a network that predicts the future latent variable sequence from the given input latent variable sequence using this latent variable sequence dataset.

The goal of the latent variable sequence generator is, given an input sequence of latent variables $(z_1, z_2, . , z_{T_{past}})$ for the past $T_{past}$ frames, to predict and generate the future latent variable sequence $(z_{T_{past}+1}, z_{T_{past}+2}, ... , z_{T_{past}+{T_{fut}}})$.
The latent variables obtained from the image reconstruction network can be roughly divided into two categories: the pose latent variable $z_{p}$ ($z^i_{where}$ and $z^i_{p}$), which varies with time, and the content latent variable $z_{c}$ ($z_{back}$ and $z^i_{c}$), which is time-independent.
Since the content latent variables are independent of time, the content latent variables in the input can also be used as the future content latent variables. Therefore, the vector $\bar{z}_{c}$ averaged over all the content latent variables in the input is used in all frames. Then, only the change in the pose latent variable is predicted.


The latent variable sequence generator predicts the future latent variable sequence using an LSTM~\cite{hochreiter1997long}. We input three vectors at each time step: the content latent variable $\bar{z}_{c}$, the pose latent variable ${z_t}_{p}$, and the difference between the previous and the current pose latent variable $d_t = {z_t}_{p} - {z_{t-1}}_{p}$ and the model estimates the predicted difference $\hat{d}_{t+1}$. The sum of the predicted difference and the current pose latent variable, ${{}\hat{z}_{t+1}}_{p} = {z_t}_{p} + \hat{d}_{t+1}$, is then used as the pose latent variable for the next time step. By repeating this process, we predict the future sequence of pose latent variables.
The loss function is the Huber loss between the predicted difference vector of the pose variables and the difference vector of Ground Truth.
In order to reduce memory usage, the training of the latent variable sequence generator does not use any type of pixel-wise reconstruction error from the videos decoded from the latent variables.
\section{Experiments for video frame reconstruction \label{sec:experiment_image}}
This section evaluates the performance of the image reconstruction network proposed in Sec.~\ref{sec:image_generator}.
We evaluate qualitatively and quantitatively whether each module of the proposed network can reconstruct the corresponding image frame, and whether the obtained latent variables are decomposed according to the hierarchical structure of the video.


For this evaluation, we used the ``Moving MNIST'' dataset, whose videos consist of $0$ to $2$ MNIST~\cite{lecun1998mnist} digits enlarged to $42 \times 42$ moving over a background from CIFAR-10~\cite{krizhevsky2009learning} enlarged to $128 \times 128$. Digits change colors over time, and their motion is reflected when hitting the edge of the image or another digit. In this dataset, the position and color of the digits are the \textit{pose information} (i.e., variant) of the foreground object, and the shape of the digits is \textit{the content information} (i.e., invariant). This dataset is much more complex than the Moving MNIST datasets used in previous works such as DRNET~\cite{denton2017unsupervised} and DDPAE~\cite{hsieh2018learning}.


Our results were compared with those of DRNET \cite{denton2017unsupervised} and MoCoGAN \cite{tulyakov2018mocogan}. We added an encoder network to MoCoGAN to enable the future prediction of a given input video.


\subsection{Qualitative evaluation}



\subsubsection{Video frame reconstruction}
\begin{figure}[t]
  \begin{center} 
    \includegraphics[width=\linewidth]{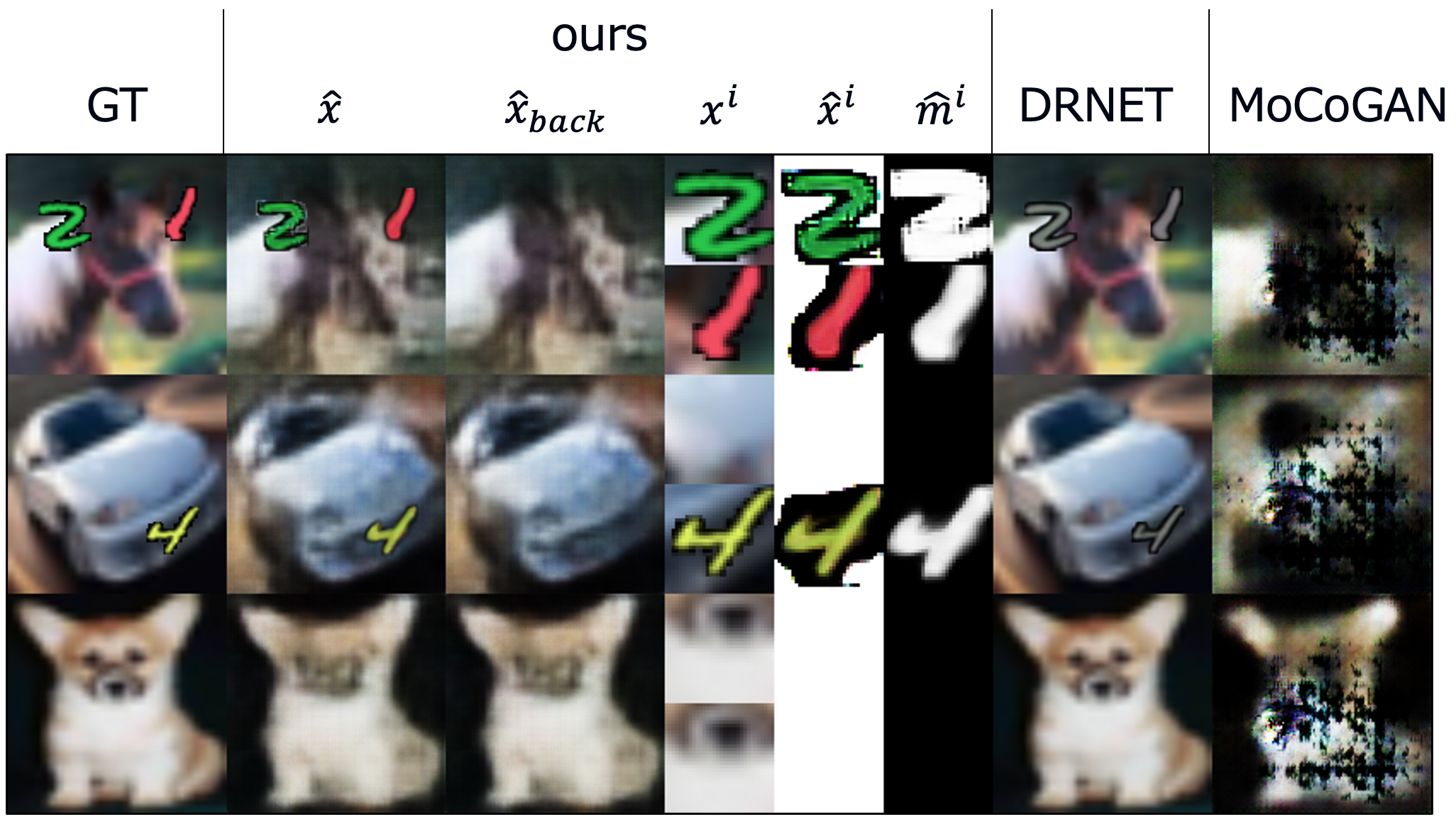}
    \caption{Video frame reconstruction results by each module of the proposed method and the comparison methods.} 
    \label{fig:image_results}
    \vspace{-5mm}
  \end{center}
\end{figure}

We qualitatively evaluate whether each module of the proposed method correctly reconstructs the background and objects respectively, and whether our method can successfully recreate the whole image.

\figref{fig:image_results} shows examples of the reconstruction by our method and the comparison methods.
Given an input frame, the background VAE can correctly reconstruct only the background, ignoring the foreground digits. It can also be seen that the object VAE correctly detects the foreground digits and outputs the corresponding images and masks. The objects are drawn at the correct position over the background to reconstruct the whole image.
When the number of objects is less than $N$, the masks for the unused masks are set to $0$ (see second and third rows in~\figref{fig:image_results}).
DRNET achieves a clean background thanks to the skip-connections of its architecture, but the colors of the foreground digits are not properly reconstructed. MoCoGAN cannot reconstruct the video frame at all.

\subsubsection{Hierarchical disentanglement}
\begin{figure}[tbp]
  \begin{center}
    \begin{tabular}{c}
      \begin{minipage}{\linewidth}
        \begin{center}
          \includegraphics[width=\linewidth]{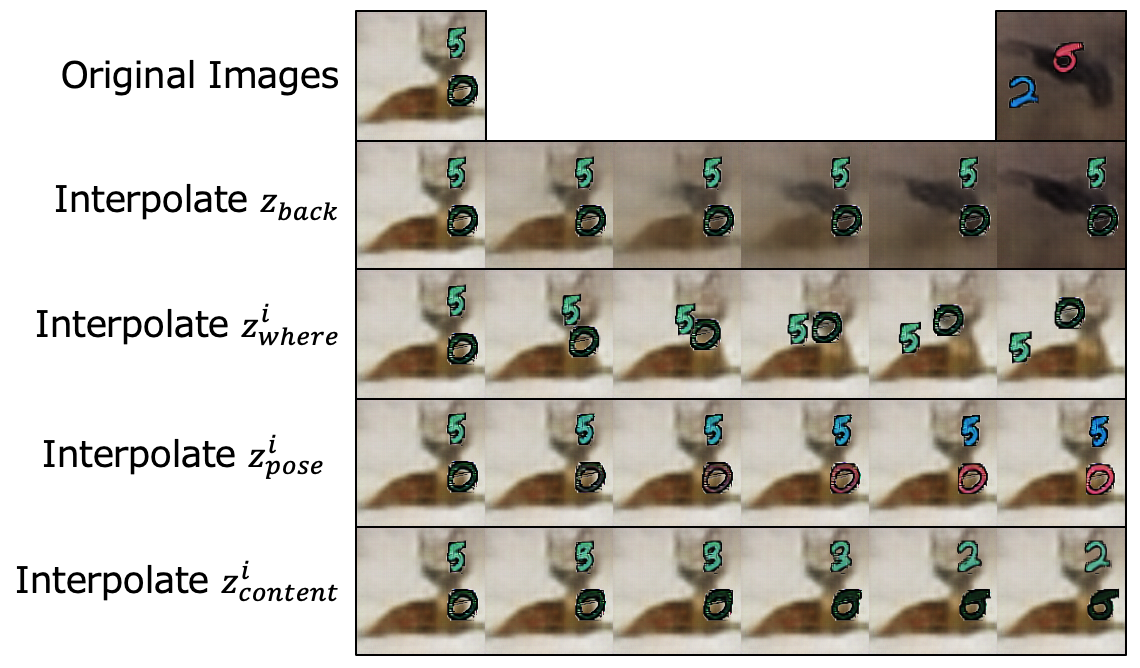}
        \end{center}
      \end{minipage} \\


    \end{tabular}
    \caption{Latent variable interpolation results. 1st row: The two original reconstructed images. 2nd to 5th row: Interpolation of each latent variable.} 
    \label{fig:interpolate}
    \vspace{-5mm}
  \end{center}
\end{figure}

Our method method disentangles video frames hierarchically into four latent variables: $z_{back}$, $z^i_{where}$,  $z^i_{p}$ and $z^i_{c}$ (i.e., background, position, pose/color, content/object type).
We encoded the four latent variables from two images $x_1$ and $x_2$, and then linearly interpolated a single latent variable between $x_1$ and $x_2$, while fixing the rest three variables.
\figref{fig:interpolate} shows that, for each latent variable, only its corresponding elements change, while the rest remain invariant. These results validate the capability of our method to disentangle the hierarchical structure of the video frames into separate latent variables.

\subsection{Quantitative evaluation (disentanglement) \label{sec:image_quantitave}}
We evaluated quantitatively the correct disentanglement of our latent variables. 
For the sake of fairness in the comparison with other methods, we consider only the decomposition into two parts: $z_{p}$ and $z_{c}$ (i.e., variant and invariant information).


In this evaluation, we train a simple network (esp. three fully connected layers) to predict certain characteristics of a video frame from the latent variables only. Then, if the latent variables are properly disentangled, only some of them will be useful for prediction, while others will be useless. For example, if we aim to predict the sum of the digits in a frame, $z_{c}$ is important since it represents the number itself, while $z_{p}$ can be ignored.
Therefore, to measure the degree of disentanglement, we use the ratio between the accuracy obtained when a single latent variable is used for training, against the accuracy of using all latent variables. A good disentanglement provides a ratio close to $1$ when the useful variables are used and a small ratio when the useless variables are used.
For this, we conducted three experiments;
(1) predicting the sum of the digits in the image, (2) predicting the class of the CIFAR-10 background image, and (3) predicting the sum of the pixel values of the digits (their color).

We also provide an ablation study of two key techniques of our method: the exchange of content latent variables (\textit{swap}) and the use of an adversarial loss for the pair of pose latent variables (\textit{adv}).

\begin{table*}[t]
    \label{tab:disentanglement}
    \caption{Qualitative results for the latent variables disentanglement. The numbers in brackets are the standard deviation among five trials.}
    \begin{center}
    \begin{tabular}{c}
    \begin{minipage}[t]{0.35\linewidth}
      \begin{tabular}{c|c|c|c}
         & $z$ & $z_{c} / z$ & $z_{p} / z$\\ \hline
        ours & \twolines{0.933}{(0.007)} & \twolines{0.994}{(0.005)} & \twolines{\bf{0.269}}{(0.008)} \\ \hline
        \twolines{ours}{(w/o adv)} & \twolines{0.936}{(0.002)} & \twolines{0.994}{(0.010)} & \twolines{0.316}{(0.070)} \\ \hline
        \twolines{ours}{(w/o swap)} & \twolines{0.837}{(0.160)} & \twolines{0.995}{(0.007)} & \twolines{0.305}{(0.063)} \\ \hline
        \twolines{ours}{(w/o adv, swap)} & \twolines{0.930}{(0.007)} & \twolines{0.981}{(0.008)} & \twolines{0.353}{(0.025)} \\ \hline
        \twolines{DRNET}{\cite{denton2017unsupervised}}  & \twolines{0.196}{(0.038)} & \twolines{\bf{1.025}}{(0.049)} & \twolines{0.854}{(0.107)} \\ \hline
        \twolines{MoCoGAN}{\cite{tulyakov2018mocogan}}  & \twolines{0.127}{(0.005)} & \twolines{1.011}{(0.034)} & \twolines{1.036}{(0.045)}
      \end{tabular}
      Accuracy on the digit sum prediction
      \label{tab:addition}
    \end{minipage}
    
    \begin{minipage}[t]{0.22\linewidth}
      \begin{tabular}{c|c|c}
        $z$ &$z_{c} / z$ & $z_{p} / z$\\ \hline
        \twolines{0.522}{(0.010)} & \twolines{1.001}{(0.022)} & \twolines{0.228}{(0.017)} \\ \hline
        \twolines{0.515}{(0.011)} & \twolines{\bf{1.050}}{(0.022)} & \twolines{\bf{0.189}}{(0.014)} \\ \hline
        \twolines{0.526}{(0.011)} & \twolines{0.985}{(0.022)} & \twolines{0.232}{(0.064)} \\ \hline
        \twolines{0.521}{(0.010)} &  \twolines{1.023}{(0.022)} & \twolines{0.202}{(0.022)} \\ \hline
        \twolines{0.102}{(0.004)} & \twolines{0.965}{(0.094)} & \twolines{0.972}{(0.113)} \\ \hline
        \twolines{0.468}{(0.011)} & \twolines{1.003}{(0.024)} & \twolines{0.574}{(0.047)}
      \end{tabular}
      Accuracy on \\ the Cifar-10 classification
      \label{tab:cifar}
    \end{minipage}
    
    \begin{minipage}[t]{0.22\linewidth}
      \begin{tabular}{c|c|c}
        $z$ & $z_{c} / z$ & $z_{p} / z$\\ \hline
        \twolines{0.00614}{(0.00097)} & \twolines{\bf{16.8}}{(2.5)} & \twolines{1.22}{(0.10)} \\ \hline
        \twolines{0.00673}{(0.00088)} & \twolines{13.4}{(1.9)} & \twolines{1.91}{(0.55)}  \\ \hline
        \twolines{0.0181}{(0.025)} & \twolines{7.88}{(3.70)} & \twolines{7.89}{(3.60)} \\ \hline
        \twolines{0.00558}{(0.00085)} & \twolines{10.2}{(3.2)} & \twolines{8.93}{(4.17)} \\ \hline
        \twolines{0.129}{(0.046)} & \twolines{1.08}{(0.10)} & \twolines{1.08}{(0.08)} \\ \hline
        \twolines{0.217}{(0.003)} & \twolines{1.01}{(0.01)} & \twolines{\bf{1.01}}{(0.01)}
      \end{tabular}
      MSE on the digits' pixel value prediction
      \label{tab:pixel}
    \end{minipage}
    \end{tabular}
    \vspace{-5mm}
    \end{center}
\end{table*}

The experimental results are shown in \tabref{tab:disentanglement}. For all three tasks, the larger $z_{c}/z$ and the smaller $z_{p}/z$ are, the better the disentanglement.
These results show quantitatively that our latent variables provide a richer representation than those of the related work, and that both \textit{adv} and \textit{swap} techniques benefit such disentangling.



\section{Experiments for future video generation}
This section evaluates experiments on the entire proposed method, which combines the image reconstruction network described in sec \ref{sec:image_generator} and the latent variable sequence generator described in sec \ref{sec:latent_generator}.

We evaluated qualitatively whether the generated future prediction video is natural, and quantitatively whether the generated future prediction video is close to the Ground Truth.
For this, two datasets are used.
The first dataset is the ``Moving MNIST'' dataset used in sec \ref{sec:experiment_image}.
The second dataset is a subset of the ``CMU motion capture'' dataset \cite{cmu}, which contains videos of subjects performing a given action in a laboratory. The videos were cropped to be square and resized to $64 \times 64$.
We used the videos of subjects with ID: 1, 2, 8, 9, 15, 17, 18, 19, 20, 21, 22, and 23. The videos from subjects between 18 to 23 are videos of two people interacting, while the remaining videos show only one person.
For both datasets, $16$ frames were taken as input and the following $16$ future frames were predicted and generated.
We compared our results with DRNET \cite{denton2017unsupervised} and MoCoGAN \cite{tulyakov2018mocogan}.

\subsection{Moving MNIST dataset}
\subsubsection{Qualitative Evaluation}
\begin{figure}[tbp]
  \begin{center}
    \begin{tabular}{c}
      \begin{minipage}{\linewidth}
        \begin{center}
          \includegraphics[width=\linewidth]{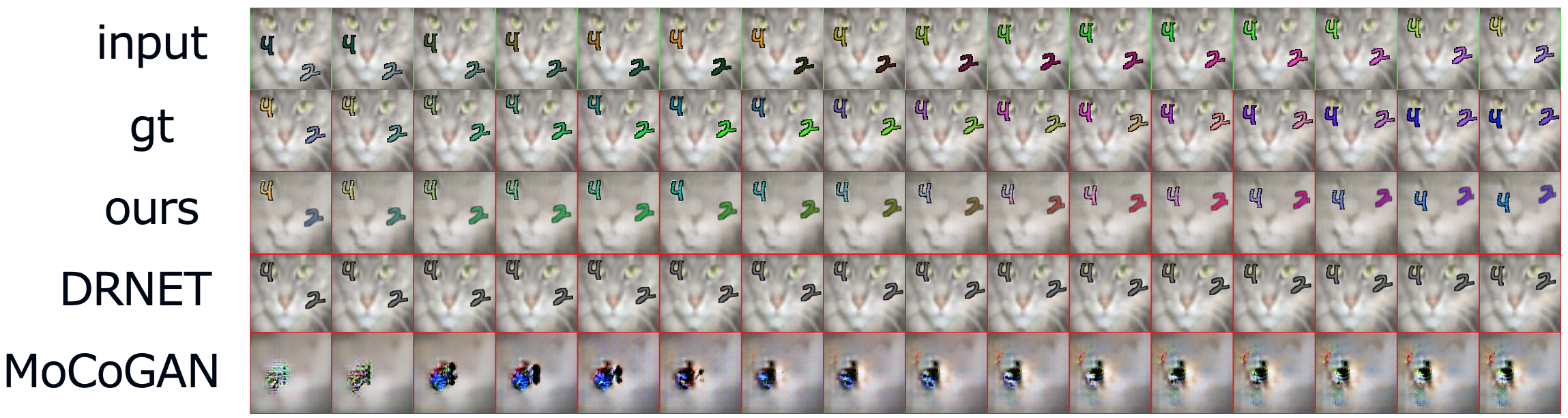}
        \end{center}
      \end{minipage} 


    \end{tabular}
    \caption{Prediction results on the ``Moving MNIST'' dataset. 1st row: Input past $16$ frames. 2nd row: Future $16$ frames (Ground Truth). 3rd to 5th row: $16$ frames predicted by each method.}
    \label{fig:results_gaming}
    \vspace{-5mm}
  \end{center}
\end{figure}

\begin{figure*}[tbp]
  \begin{center}
    \begin{tabular}{c}
      \begin{minipage}{\linewidth}
        \begin{center}
          \includegraphics[width=\linewidth]{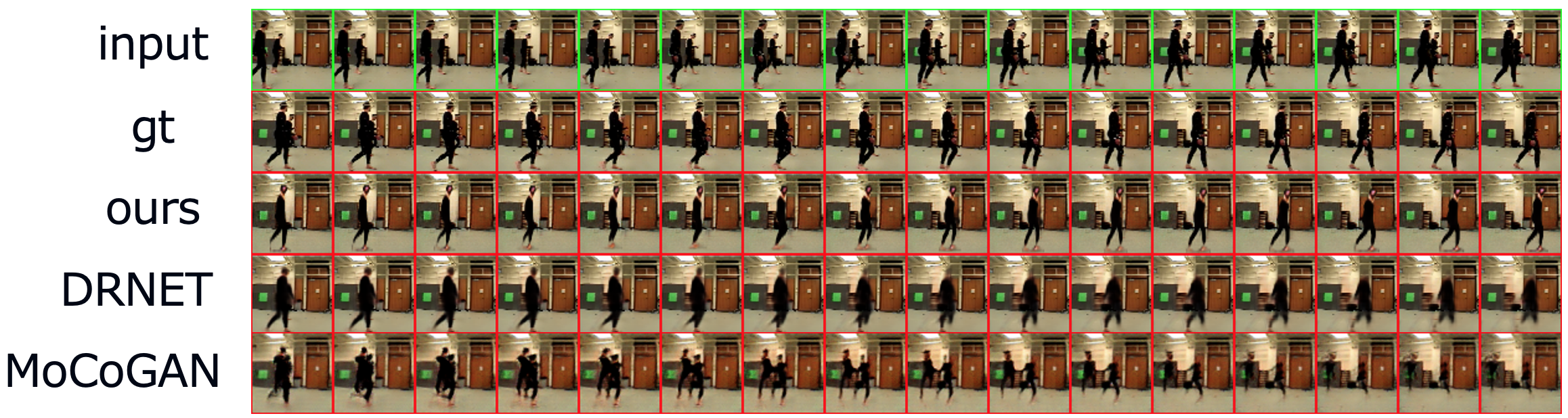}
        \end{center}
      \end{minipage} 


    \end{tabular}
    \caption{Prediction results on the CMU dataset. 1st row: Input past $16$ frames. 2nd row: Future $16$ frames (Ground Truth). 3rd to 5th row: $16$ frames predicted by each method.}
    \label{fig:results_cmu}
    \vspace{-5mm}
  \end{center}
\end{figure*}

\figref{fig:results_gaming} shows the qualitative evaluation of our method on the Moving MNIST dataset (additional results are shown in the appendix). The proposed method is able to generate a natural future prediction video, although the background is slightly blurred. In addition, the shapes of the digits remain accurate, and their predicted position and color are close to the Ground Truth.
DRNET is able to reconstruct a sharp-looking background due to its skip-connections, but it is not able to learn the color of the foreground digits.
Finally, MoCoGAN performance is not accurate at all.

\subsubsection{Quantitative Evaluation}

We evaluated how close the generated future video is to the Ground Truth.
Since a pixel-wise distance-based measure with respect to the Ground Truth would focus on the background and practically ignore the foreground objects, we chose three different metrics. First, we pre-trained a network using VGG16~\cite{Simonyan15} for three prediction tasks: the sum of the positions of the foreground digits, the sum of the pixel values (color), and the sum of the digits themselves in the generated video. For each task, our metric is how close the output of the network is to the ground truth label given the generated frame.

\begin{figure}[t]
    \centering
    \includegraphics[width=\linewidth]{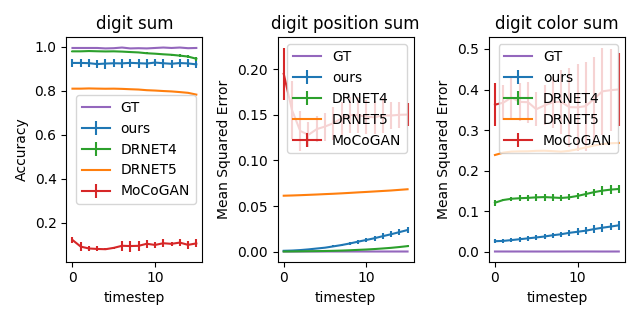}
    \caption{Evaluation of the generated video on the Moving MNIST dataset. Left: Accuracy of the sum of the digits. Center: MSE of the sum of the positions of the digits. Right: MSE of the sum of the pixel values of the digits. Note that DRNET shows the average of four times, excluding the times when learning failed, and the average of five times, including the times when learning failed.}
    \label{fig:eval_gaming}
    \vspace{-3mm}
\end{figure}

\figref{fig:eval_gaming} shows the aforementioned metrics along the predicted future frames. Each method was trained 5 times and the mean and standard deviation were taken. However, since DRNET failed to train once, two graphs are shown, one with and one without the failed training.
Our method significantly outperforms the other methods in the task of predicting pixel values. On the other hand, the proposed method is inferior to DRNET in the digits sum and digits location tasks when learning did not fail. A possible reason is that, unlike DRNET whose background is less blurry (\figref{fig:results_gaming}) thanks to its skip connections, VGG16 could not cope with the domain gap between the training data and the generated data for the proposed method. However, in the digit summation task, the performance of DRNET degrades with time, while ours is consistent regardless of time. This is strong proof that our disentangled latent variables are effective for video generation.

\subsection{CMU motion capture dataset}



The goal of this experiment is to verify whether the proposed method can be applied not only to artificial videos but also to real videos. Since the number of videos in this dataset is small and quantitative evaluation is difficult, we evaluated it only qualitatively.

\figref{fig:results_cmu} shows that the foreground human in the related work methods gets gradually more and more blurry, while the proposed method can generate a clear human consistently (additional results are shown in the appendix).




\section{Conclusion}

In this paper, we proposed a two-stage method to generate a future prediction video without using the entire video for training at once to reduce memory usage. To facilitate the learning process, we proposed an image reconstruction network that can obtain latent variables disentangled according to the hierarchical structure of the video; a video can be decomposed into the background and foreground objects, and each foreground object has time-varying information and time-independent information. Our method not only is more efficient, but also the quality of our future videos is the most consistent over time.

As a future work, we plan to apply our framework to other fields besides the future video generation, such as unconditional video generation and video interpolation. It has also a potential for other video-related tasks not limited to generation, such as object detection, action recognition, and video retrieval.

\subsection*{Acknowledgements}
This work was supported by JST CREST Grant Number JPMJCR2015, and Moonshot R\&D Grant Number JPMJPS2011.

{\small
\bibliographystyle{ieee_fullname}
\bibliography{main_cameraready}
}

\clearpage
\section*{Supplementary material}
\setcounter{section}{0}
\renewcommand{\thesection}{\Alph{section}}

\section{Details of the proposed method}
\subsection{How to select $\alpha$}
This section describes how to select $\alpha$ in \equref{eq:decoder}.

In DRNET \cite{denton2017unsupervised}, the content latent variables of the same object in two different frames are completely exchanged (i.e., $\alpha=1$ always). However, unlike DRNET, our method further decomposes each object for a finer hierarchical disentanglement. We found that if we set $\alpha=1$, the learning is not stable and the model tends to fall into a local minimum where no object is detected. So, we sampled $\alpha$ from a uniform distribution of $[0, \frac{e}{E}]$, where $E$ is the total number of epochs of the learning and $e$ is the current epoch. Thus, at the beginning of the learning, $\alpha$ is always $0$ and there is no exchange of content latent variables, and as the learning progresses, the ratio of content latent variables from different frames is increased. At the beginning of the learning process, by not mixing the content latent variables, the reconstruction of the object is easier than when mixing is performed, and the learning process is stabilized to avoid falling into the local minimum where no object is detected.

\subsection{Conversion from Video Dataset to Latent Variable Sequence Dataset \label{sec:dataset_converter}}
This section describes the process of converting a video dataset into a latent variable sequence dataset using the encoders trained in Sec.~\ref{sec:image_generator}. This process is not performed multiple times during training, but only once after the image reconstruction network has been trained.

A video of duration $T$ frames is represented as $(x_1, x_2, ..., x_T)$.
As explained in Sec.~\ref{sec:merge}, given an image $x$, the latent variable $z$ corresponding to the entire image can be obtained by concatenating the latent variable $z_{back}$ (obtained from the background VAE) and the latent variable $z^i (1 \le i \le N)$ for each object (obtained from the object VAE). So, by obtaining the corresponding latent variable $z_t$ for each frame $x_t (1\le t\le T)$ of the video, and concatenating them in the time direction, we can obtain the sequence of latent variables corresponding to the video.
However, in order to obtain a latent variable sequence that is easy to learn for the latent variable sequence generator to be trained later, we need to make some modifications.


Since each $x_t$ is encoded independently, the order in which objects are detected can be different for each frame. In other words, the first object detected in the $t$ frame may be the second object detected in the $t+1$ frame. Therefore, if we just concatenate the latent variables $z_t^i$ without paying attention to the order, the latent variables do not change smoothly between adjacent frames, and it is difficult for the latent variable sequence generator to carry out predictions. Therefore, we need to align the order in which the objects are concatenated regardless of the frame.

As shown in \equref{eq:search_nearest}, the distance between the content latent variables can be used to identify the latent variables that represent the same object in different frames. When the maximum number of objects in a video is $N$, there are $N!$ different ways in which objects can be sorted. We calculate the sum of the distances of the content latent variables between the corresponding objects of adjacent frames, for all the $N!$ ways of sorting the objects. Therefore, the sorting with the smallest sum can be used to obtain the latent variables for which the order of the objects do not change between frames. The equation is as follows:

\begin{eqnarray}
    perm = \argmin_{perm} \sum_{i=1}^N \|{z_t}_{content}^i - {z_{t+1}}_{content}^{perm_i}\|^2 \\
    {z_{t+1}}^i \coloneqq {z_{t+1}}^{perm_i}
\end{eqnarray}

Where $perm$ is the permutation of the array $(1, 2, ..., N)$ that produces the desired sorting, and represents the order of the objects.

\clearpage
\onecolumn
\section{Additional Generated Video}
\subsection{Moving MNIST dataset}
\begin{figure}[h]
  \begin{center}
    \begin{tabular}{c}
      \begin{minipage}{\linewidth}
        \begin{center}
          \includegraphics[width=\linewidth]{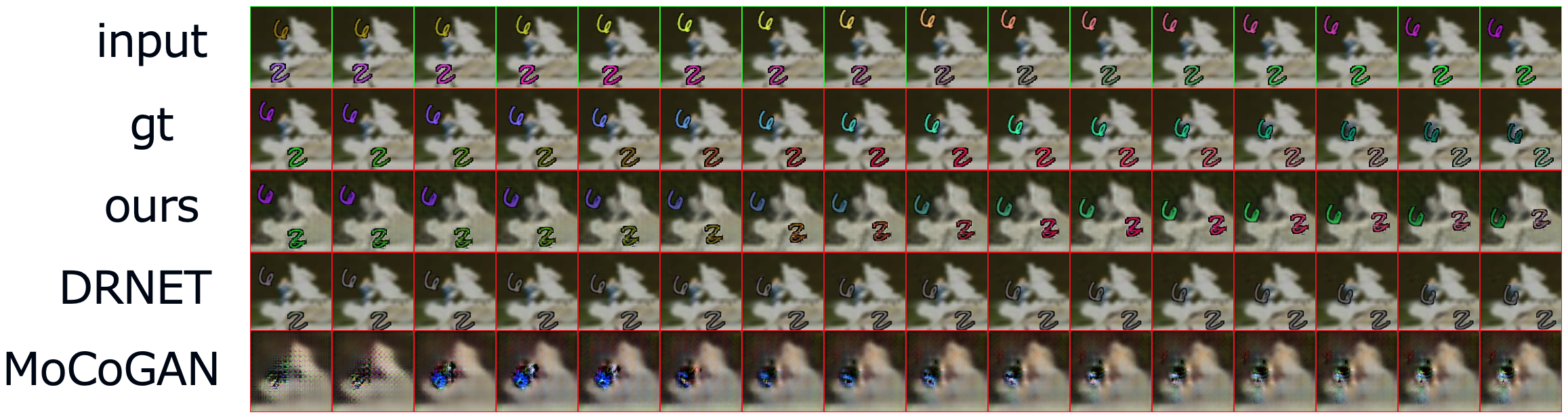}
        \end{center}
      \end{minipage} \\

      \begin{minipage}{\linewidth}
        \begin{center}
          \includegraphics[width=\linewidth]{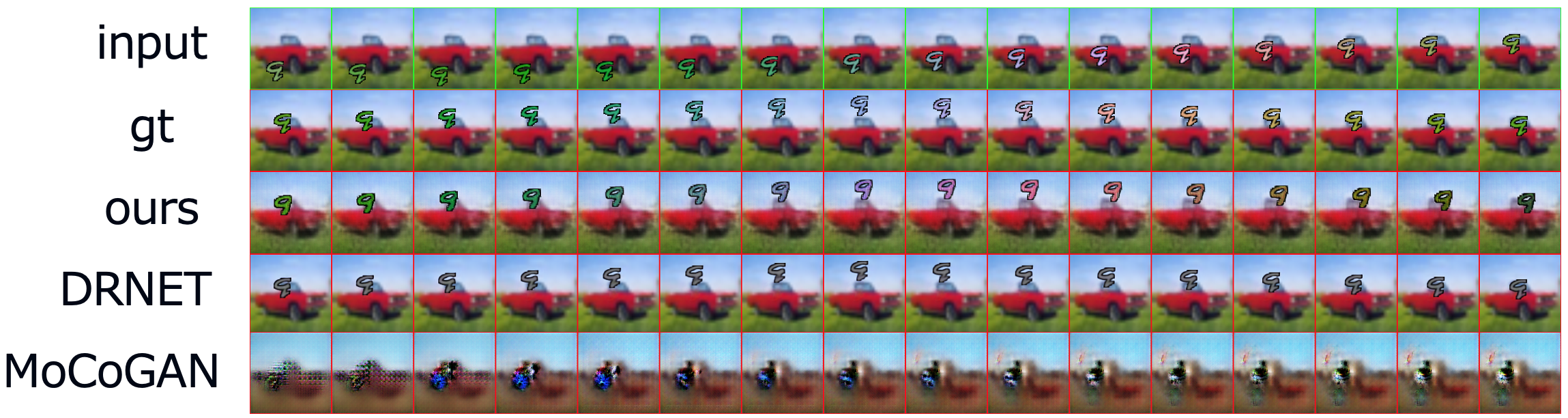}
        \end{center}
      \end{minipage}\\

      \begin{minipage}{\linewidth}
        \begin{center}
          \includegraphics[width=\linewidth]{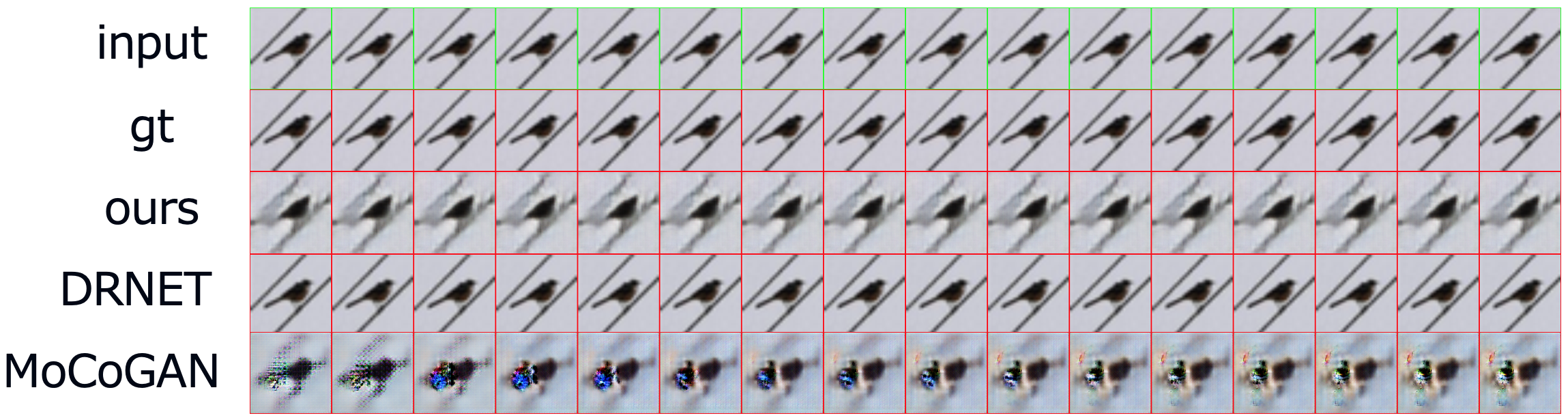}
        \end{center}
      \end{minipage}\\
    \end{tabular}
    \caption{Prediction results on the Moving MNIST dataset. 1st row: Input past $16$ frames. 2nd row: Future $16$ frames (Ground Truth). 3rd to 5th row: $16$ frames predicted by each method.}
  \end{center}
\end{figure}

\clearpage
\subsection{CMU motion capture dataset}
\begin{figure*}[h]
  \begin{center}
    \begin{tabular}{c}
      \begin{minipage}{\linewidth}
        \begin{center}
          \includegraphics[width=\linewidth]{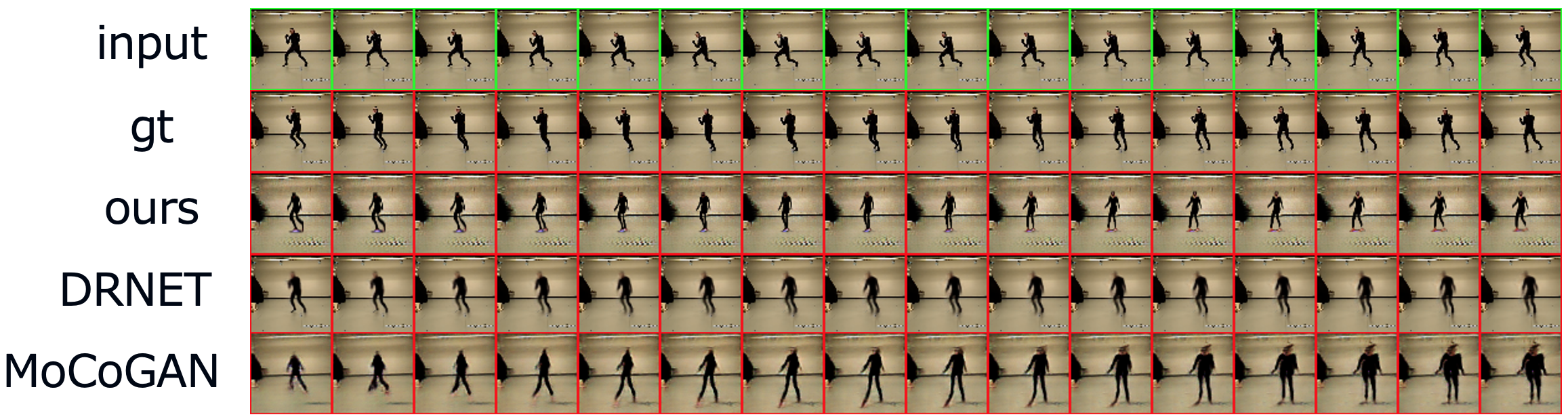}
        \end{center}
      \end{minipage} \\

      \begin{minipage}{\linewidth}
        \begin{center}
          \includegraphics[width=\linewidth]{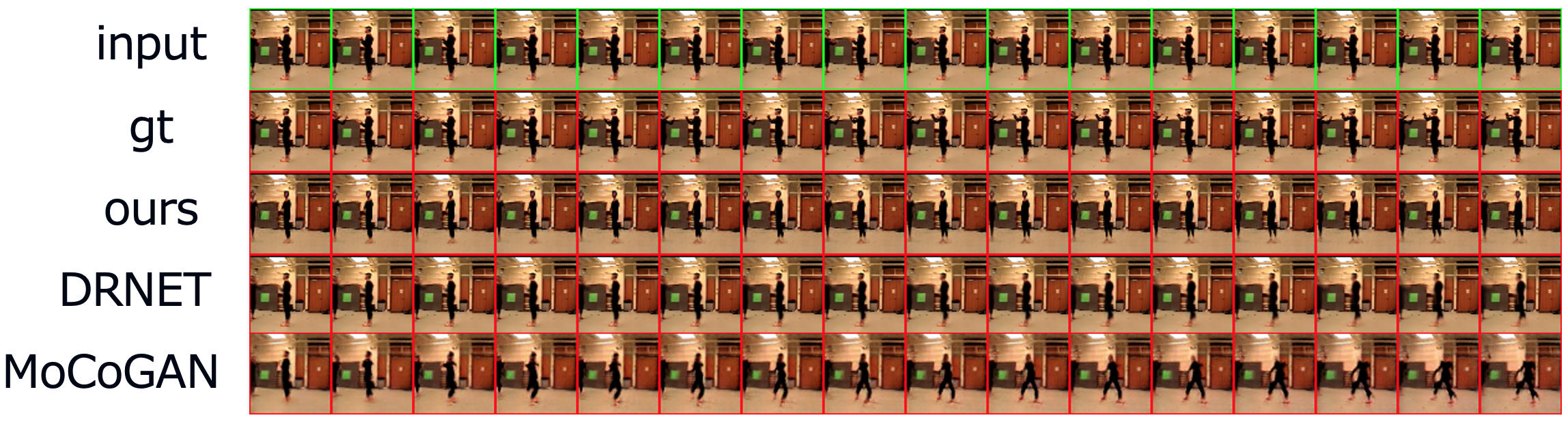}
        \end{center}
      \end{minipage}\\

      \begin{minipage}{\linewidth}
        \begin{center}
          \includegraphics[width=\linewidth]{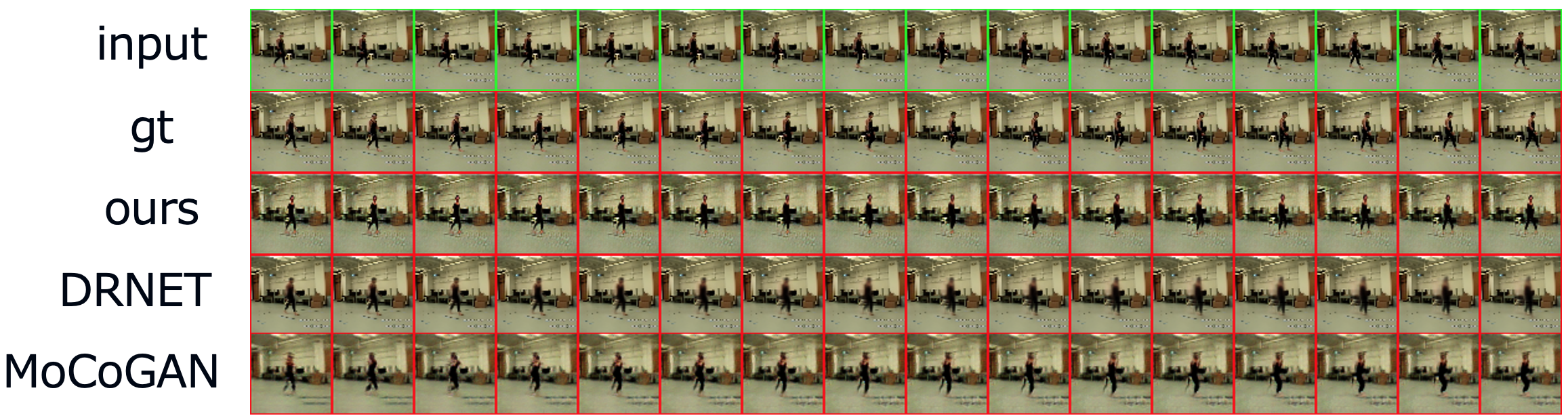}
        \end{center}
      \end{minipage}\\
    \end{tabular}
    \caption{Prediction results on the CMU dataset. 1st row: Input past $16$ frames. 2nd row: Future $16$ frames (Ground Truth). 3rd to 5th row: $16$ frames predicted by each method.}
  \end{center}
\end{figure*}

\end{document}